\documentclass{article}

\usepackage{hyperref}

\makeatletter
\def\input@path{{sty/}}
\makeatother

\usepackage{arxiv}
\usepackage{wrapfig}
\usepackage{amsmath,amssymb,amsfonts}
\usepackage{booktabs}
\usepackage{pifont}
\usepackage{placeins}

\icmltitlerunning{NeuroCanvas: VLLM-Powered Robust Seizure Detection by Reformulating Multichannel EEG as Vision}

\begin{document}

\icmltitle{NeuroCanvas: VLLM-Powered Robust Seizure Detection by Reformulating Multichannel EEG as Image
% Reformulating Multichannel EEG as Visual Intensity Maps for Robust Seizure Detection
}

% Alternative: Intensity-Based Visual Reprogramming of EEG for Seizure Detection using Vision-Language Model*



\icmlsetsymbol{equal}{*}

\begin{icmlauthorlist}
  \icmlauthor{Yan Chen}{equal,ul}
  \icmlauthor{Jie Peng}{equal,unc}
  \icmlauthor{Moajjem Hossain Chowdhury}{equal,ul}
  \icmlauthor{Tianlong Chen}{unc}
  \icmlauthor{Yunmei Liu}{ul}
\end{icmlauthorlist}

\icmlaffiliation{ul}{MINDxAI Lab, Industrial \& Systems Engineering Department, University of Louisville}
\icmlaffiliation{unc}{UNITES Lab, University of North Carolina at Chapel Hill}

% \icmlcorrespondingauthor{Yunmei Liu}{yunmei.liu@louisville.edu}

\icmlkeywords{Machine Learning, ICML}

\vskip 0.3in


\printAffiliationsAndNotice{}  % no special notice (required even if empty)

\begin{abstract}
  % Seizure is one of the most prevalent neurological conditions and the development of a robust seizure detection model is crucial for clinical intervention. Electroencephalography (EEG) is the most widely used signal for seizure detection. 
  % However, EEG-based seizure detection models remain difficult to deploy in clinical settings because seizure events are temporally sparse and frequently exhibit missing channels or corrupted segments. 
  Accurate and timely seizure detection from Electroencephalography (EEG) is critical for clinical intervention, yet manual review of long-term recordings is labor-intensive. 
  Recent efforts to encode EEG signals into large language models (LLMs) show promise in handling neural signals across diverse patients, but two significant challenges remain: (1) \textbf{multi-channel heterogeneity}, as seizure-relevant information varies substantially across EEG channels, and (2) \textbf{computing inefficiency}, as the EEG signals need to be encoded into a massive number of tokens for the prediction.
  To address these issues, we draw the EEG signal and propose the novel \textbf{NeuroCanvas} framework.
  Specifically, \textbf{NeuroCanvas} consists of two modules: (i) The \textit{Entropy-guided Channel Selector} (\texttt{ECS}) selects the seizure-relevant channels input to LLM and (ii) the following \textit{Canvas of Neuron Signal} (\texttt{CNS}) converts selected multi-channel heterogeneous EEG signals into structured visual representations.
  The \texttt{ECS} module alleviates the multi-channel heterogeneity issue, and the \texttt{CNS} uses compact visual tokens to represent the EEG signals that improve the computing efficiency.
  % We demonstrate the effectiveness of \textbf{NeuroCanvas} in seizure detection across $x$ datasets with $xxx$ F1 score improvement and $xxx$ Flops/Latency reduction.
  We evaluate NeuroCanvas across multiple seizure detection datasets, demonstrating a significant improvement of $20\%$ in F1 score and reductions of $88\%$ in inference latency. 
  These results highlight NeuroCanvas as a scalable and effective solution for real-time and resource-efficient seizure detection in clinical practice.
  The code will be released at \url{https://github.com/Yanchen30247/NeuroCanvas.git}.

\end{abstract}

\section{Introduction}

As one of the most prevalent neurological conditions, epilepsy poses a significant public health challenge \cite{singh2020global}. It is characterized by the predisposition to generate recurrent seizures and impacts the lives of approximately 50 million people worldwide \cite{WHO_Epilepsy_2024}. Consequently, accurate and timely seizure detection is vital for immediate medical intervention \cite{devinsky2016sudden}. 
Scalp electroencephalography (EEG) plays an important role in seizure detection. Clinically, the standard for identifying seizure activity involves the visual interpretation of long-term EEG recordings by specialized physicians or neurologists. However, this manual review process is labor-intensive and time-consuming \cite{ramgopal2014seizure}.

To address these limitations, automated seizure detection systems based on deep learning have been proposed. These systems leverage advanced algorithms to enable continuous monitoring without the constraints of human fatigue \cite{rasheed2020machine}. For instance, \citet{tang2021self} employed self-supervised graph neural networks (GNNs), while \citet{afzal2024rest} utilized efficient recurrent update mechanisms to achieve impressive detection accuracy. Latest Evobrain \cite{kotoge2025evobrain} further integrated a two-stream architecture within a time-graph framework to explicitly model the dynamic evolution of brain networks. Despite these advancements, two critical challenges persist: (1) seizure events are temporally sparse, occurring far less frequently than non-seizure events, and (2) EEG signals vary significantly across patients, limiting the generalization of deep learning models in practical scenarios.

Recently, large language models (LLMs) have shown remarkable potential in addressing these challenges due to their strong generalization capabilities derived from pretraining. Pioneering works like NeuroLM~\cite{jiang2024neurolm}, UniMind~\cite{lu2025unimind}, and EEG-GPT~\cite{kim2024eeg} have explored novel approaches to tokenizing EEG signals or translating EEG features into verbal representations, enabling chat-style LLMs to perform prediction and detection tasks. 
These methods demonstrate promise in leveraging LLMs for seizure detection by handling both the scarcity of seizure events and patient-specific signal differences.
However, integrating LLMs for EEG-based seizure detection is not without challenges. Two critical issues remain unresolved: (1) \textbf{Multi-channel heterogeneity}, where the importance of different EEG channels varies significantly across patients and environments, and indiscriminate use of multi-channel signals may introduce irrelevant noise that decrease prediction accuracy.
(2) \textbf{Computing inefficiency}, as existing methods require massive tokenization of multi-channel signals, resulting in excessive computational demands that are incompatible with real-time seizure detection requirements. 
Addressing these challenges is crucial for developing scalable, efficient, and accurate seizure detection systems for practical clinical settings.

Motivated by the observations above and the limitations in seizure detection, we propose \texttt{NeuroCanvas}, a novel framework that innovatively transforms EEG signals into visual representations, enabling robust seizure detection using vision-based large language models (VLLMs). Specifically, NeuroCanvas comprises two key components: ($i$) The Entropy-guided Channel Selector (\texttt{ECS}): This module uses channel-wise entropy to rank electrodes based on their informativeness, retaining only the most relevant channels for seizure detection.
By filtering out irrelevant channels, the \texttt{ECS} module reduces noise and improves task accuracy, addressing the issue of multi-channel heterogeneity.
($ii$) The \texttt{CNS}: This module transforms multi-channel EEG windows into compact intensity maps, where pixel values encode normalized signal activity. Seizure-related bursts appear as salient spatiotemporal motifs, which can be effectively captured by pretrained visual models. The \texttt{CNS} module enhances computational efficiency by reducing the input tokens while preserving critical information.
Finally, we fine-tune a pretrained VLLM to adapt its general-purpose visual representations for accurate seizure detection. 
% Fine-tuning leverages prompt-guided mechanisms tailored for clinical EEG datasets, enabling reliable detection.
In summary, our contributions are as follows:
\begin{itemize}
    \item \textbf{EEG-to-Image Encoding:} We introduce an innovative strategy to encode multi-channel EEG signals into compact intensity maps, enabling pretrained visual models to effectively capture seizure-related patterns.
    \item \textbf{VLLM-Driven Framework:} We propose a robust seizure detection framework that combines entropy-guided channel selection with prompt-guided VLLM adaptation, ensuring efficient inference and maintaining reliability by filtering irrelevant EEG channels.
    \item \textbf{Experimental Validation:} Extensive experiments on the TUSZ and CHB-MIT datasets demonstrate the superiority of NeuroCanvas, achieving a binary F1-score of $0.501$, over $20\%$ higher than the best previous model. Furthermore, NeuroCanvas remains robust under severe channel reduction, achieving accuracy of $0.8487$ with only two EEG channels retained, and provides low-latency inference with $19$ms per sample.
\end{itemize}

\section{Related Work}

\textbf{Deep Learning Models for EEG Seizure Detection.}
Deep learning have become the dominant method for seizure detection \cite{rasheed2020machine}. Classical methods widely used Convolutional Neural Networks (CNN) and Recurrent Neural Networks (RNN) to capture spatiotemporal dependencies in EEG signals \cite{acharya2018deep,emami2019seizure,o2020neonatal,talathi2017deep,tsiouris2018long,zhang2022epileptic}. To further capture the non-Euclidean nature of brain connectivity, GNNs, such as DCRNN \cite{tang2021self}, REST \cite{afzal2024rest} and EvoBrain \cite{kotoge2025evobrain}, were proposed to model topological relationships. However, these models struggle with fixed channel configurations and poor generalization ability. Despite the architectural advancements of GNN-based models, they remain constrained by fixed channel configurations and poor generalization ability \cite{afzal2024rest,tang2021self}. Thus, their performance decreases in clinical EEG signals characterized by heterogeneity and incompleteness.

% Convolutional Neural Networks (CNN) are widely adopted to capture spectral and spatial features by transforming raw EEG signals into spectrograms or treating multi-channel recordings as Euclidean 2-dimension images \cite{acharya2018deep,emami2019seizure,o2020neonatal,fan2024domain}. Although CNN-based models are efficient in feature extraction, they often lack specific mechanisms to handle the sequential nature of physiological data. To address this, Recurrent Neural Networks (RNN), particularly Long Short-Term Memory (LSTM) networks, were employed to model the long-term temporal dependencies inherent in EEG signals\cite{talathi2017deep,tsiouris2018long,zhang2022epileptic}. 

% However, both traditional CNN and RNN typically treat EEG electrodes as a regular grid or independent channels, resulting in discarding the non-Euclidean geometry of brain connectivity \cite{tang2021self,ho2023self}. To bridge this gap, Graph Neural Networks (GNNs) have been proposed to model the topological relationships between electrodes. For instance, \citet{tang2021self} introduced the Diffusion Convolutional Recurrent Neural Network (DCRNN), which integrates graph diffusion processes to capture dynamic spatiotemporal correlations across the brain network. Following this approach, \citet{afzal2024rest} proposed REST, a graph-based framework utilizing residual state updates to address the efficiency bottlenecks of previous graph models.
\textbf{EEG–LLM Alignment and Tokenization.}
% Recently, the rapid development of foundations models raises the potential to solve these limitations in the seizure detection. These approaches typically align EEG signals with the semantic space of LLMs to leverage their robust generalization capabilities. For instance, NeuroLM treats EEG signals as a "foreign language," employing a vector-quantized tokenizer to encode continuous signals into discrete neural tokens for LLM processing \cite{jiang2024neurolm}. Similarly, \citet{lu2025unimind} introduced UniMind, which utilizes a neuro-language connector to transform complex spatiotemporal EEG patterns into compact neuro-semantic tokens.
Recently, the rapid development of foundations models raises the potential to solve these limitations in the seizure detection. These approaches typically align EEG signals with the semantic space of LLMs to leverage their robust generalization capabilities \cite{jiang2024neurolm, lu2025unimind}.
% Add evobrain here， compare our model with theirs
However, application of text-based LLMs to high-dimensional EEG signals presents critical bottlenecks: token inefficiency and morphological information loss. As EEG is continuous and high dimensional, tokenization often yields long sequences that consume a disproportionate amount of the LLM's context window \cite{liu2025picture,merrill2024language}. Moreover, discrete tokens often fail to preserve fine-grained morphological features, such as the precise geometry of spikes or subtle rhythm changes, which are more naturally preserved in visual representations \cite{zhang2025timemaster,liu2025picture}.
VLLM based approaches, thus, represent a promising paradigm which takes advantage of the inherent generalization of VLLM foundation models while preserving the morphological integrity\cite{zeng2025wavemind,liu2025picture}.
\textbf{Vision-Based Encoding for Time-Series.}
% Classical visual representation of EEG is time-frequency transforms such as the short-time Fourier transform (STFT) or Continuous Wavelet Transform (CWT) \cite{peng2022seizure,li2018epileptic,faust2015wavelet}. While this representation effectively renders non-stationary neural oscillations as distinct visual textures accessible to models like CNN, it faces severe scalability issues in clinical settings. Specifically, since STFT or CWT operate on a single-channel basis, processing a multi-channel EEG recording requires generating independent high-resolution two-dimensional images for every channel. Concatenating these spectrograms results in a massive input tensor with excessive spatial redundancy, which significantly degrades token efficiency when processed by Foundation Models with fixed context windows \cite{liu2025picture,allen2010time}. Moreover, the high computational complexity required to compute time-frequency decompositions for multiple channels causes substantial latency, thereby obstructing the deployment of such systems for real-time seizure prediction \cite{allen2010time}.
To align physiological data with the pre-trained VLLMs, recent frameworks have explored directly rendering time-series as waveform line plots with different color representing different channels \citet{zhang2025timemaster, liu2025picture,he2025harnessing}. However, such images are dominated by background, which makes Vision Transformers to inefficiently allocate attention resources to redundant pixels rather than features \cite{liu2024revisiting,marchetti2025efficient} 
% To align physiological data with the pre-trained VLLMs, recent frameworks have explored directly rendering time-series as waveform line plots. For instance, \citet{zhang2025timemaster} proposed TimeMaster, which visualizes time-series data as multi-channel line graphs, using distinct colors to represent different channels for LLM reasoning. However, this representation suffers from low information density and visual ambiguity. Such images are dominated by background, which makes Vision Transformers to inefficiently allocate attention resources to redundant pixels rather than features \cite{liu2024revisiting,marchetti2025efficient}. Moreover, since EEG signals contain many channels and channels typically share similar amplitude ranges, projecting multiple channels into same two-dimensional coordinate system often results in severe signal overlap and occlusion.
% @Jie does this weaken our contribution?
Distinct from sparse waveform plots, dense encoding strategies map normalized signal amplitudes directly into pixel intensity grids \cite{ni2025harnessing, chen2024visionts}. Such representations preserve the intrinsic time–channel topology while reducing non-informative background regions, resulting in more compact visual inputs. Under fixed visual token budgets, this increased information density enables more efficient use of visual patches by reducing attention allocated to background regions \cite{endo2025feather}.

\section{Background}

\textbf{Characteristics of EEG Seizure Data.}
In clinical environments, seizure detection is characterized by extreme class imbalance. Previous deep learning methods typically employ the balancing strategy to artificially equilibrate the training set \cite{zhang2022epileptic,tang2021self,afzal2024rest}. However, in previous deep learning models, this strategy introduces a prior shift between training and inference which biases the model toward the positive class. Consequently, when deployed for test recordings where non-seizure segments are overwhelmingly dominant, such models are prone to excessive false positive rates\cite{lipton2018detecting,ingolfsson2024minimizing}. Compounding this issue is the challenge of channel heterogeneity as clinical EEG recordings frequently feature inconsistent or missing channels as seen in EEG datasets \cite{shah2018temple, guttag2010chb}. 

\textbf{Visual Representations of EEG Signals.}
Classical visual representation of EEG is time-frequency transforms such as the short-time Fourier transform (STFT) or Continuous Wavelet Transform (CWT) \cite{peng2022seizure,li2018epileptic,faust2015wavelet}. While this representation effectively renders non-stationary neural oscillations as distinct visual textures accessible to models like CNN, it faces severe scalability issues in clinical settings. Specifically, since STFT or CWT operate on a single-channel basis, processing a multi-channel EEG recording requires generating independent high-resolution two-dimensional images for every channel. Concatenating these spectrograms results in a massive input tensor with excessive spatial redundancy \cite{allen2010time}. Moreover, the high computational complexity required to compute time-frequency decompositions for multiple channels causes substantial latency, thereby obstructing the deployment of such systems for real-time seizure prediction \cite{allen2010time}.

% \subsection{Foundation Models for Physiological Time-Series}

% \begin{wrapfigure}[18]{l}{0.46\columnwidth}
% \captionsetup{skip=3pt}
% \includegraphics[width=\linewidth,height=0.28\textheight,keepaspectratio]{fig1_overview.pdf}
% \caption{Overview of the NeuroCanvas framework.}
% \label{fig:fr_overview}
% \end{figure}

% We adopt our \textbf{NeuroCanvas} framework on the seizure detection task to show the advance.
% Specifically, we define the seizure detection task, formally, and introcuce the framework overview.
% Then, we detailed introduce the \textit{Entropy-guided Channel Selector} (\texttt{ECS}) module.
% At last, we put the selected EEG channels to the canvas and explain our novel \textit{Canvas of Neuron Signal} (\texttt{CNS}) module.

% We apply our \textbf{NeuroCanvas} framework to the seizure detection task to demonstrate its advantages.
This work introduces \textbf{NeuroCanvas}, a novel EEG representation that maps multichannel EEG signals into information-dense intensity grids, enabling efficient and morphology-preserving alignment with vision–language models.
We begin by formally defining the seizure detection task and providing an overview of the framework (Section~\ref{problem_form_overview}). 
Next, we introduce the \textit{Entropy-guided Channel Selector} (\texttt{ECS}) module, which identifies the most informative EEG channels for the task based on their entropy (Section~\ref{ecs_module}). 
Finally, we present our novel \texttt{CNS} module, which visualizes the selected EEG channels on the ``canvas'' for vision large language mdoel training (Section~\ref{cns_module}). Detailed description about the base model architecture is in appendix \ref{moedl_arc}

\subsection{Problem Formulation and Framework Overview}\label{problem_form_overview}

\textbf{Problem formulation.} 
We formulate the seizure detection task as a binary classification problem over multi-channel EEG signals. $\mathcal{D} = \{(X^{(i)}, y^{(i)})\}_{i=1}^N$ represents a dataset that includes $N$ EEG segments. Each input $X^{(i)} \in \mathbb{R} ^ {C_i \times T}$ represents a EEG clip with $C_i$ channels and $T$ time steps. The corresponding label $y^{(i)} \in \{0,1\}$ indicates the annotation, where $y^{(i)} = 1$ donates the presence of seizure event and $y^{(i)} = 0$ donates the normal background (non-seizure). 

\begin{wrapfigure}[28]{r}{0.51\textwidth}
  \centering
  \includegraphics[width=\linewidth]{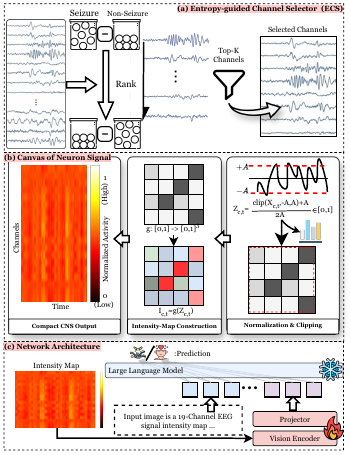}
  \caption{Overview of the NeuroCanvas framework.}
  \label{fig:fr_overview}
\end{wrapfigure}

\textbf{Framework overview.} The proposed NeuroCanvas operates in three parts. 
\textit{(a)} Raw EEG signals are first received by the \texttt{ECS} module. Spectral entropy is calculated to find the most discriminative EEG channels for seizure detection. The top $K$ channels selected by \texttt{ECS} is then fed to the \texttt{CNS} module(Figure \ref{fig:fr_overview}(a)). 
\textit{(b)} The \texttt{CNS} module functions as a universal adapter for heterogeneous EEG signals. The selected signals are first normalized and clipped then they are mapped and chromatically encoded to generate a "Visual Canvas", donated as $I_{\text{c}} \in \mathbb{R}^{H \times W \times 3}$, where $H$ and $W$ represent the resolution and $3$ corresponds to the RGB color channels (Figure \ref{fig:fr_overview}(b)).
\textit{(c)} Subsequently, the generated "Visual Canvas" will be input to a pretrained VLLM for seizure prediction (Figure \ref{fig:fr_overview}(c)). In this stage, a visual encoder $E_{\text{v}}$ first extract spatial embeddings $Z_v = E_{\text{v}}(I_{\text{c}})$. These visual embeddings are concatenated with the token embeddings of a text prompt. Eventually, the LLM base model $\mathcal{M}$ predicts the seizure detection result.

% After preprocessing and clipping, we get EEG signal clips $X$ and labels $y$ for the detection task. For each EEG clip $X \in \mathbb{R}^{C \times T}$, $C$ represents the number of channels and $R$ represents the time steps. In detection task, the label is binary, where 0 represents for non-seizure and 1 represents for seizure.

\subsection{Entropy-guided Channel Selection}\label{ecs_module}
% This part is about the channel selection method. Include the discussion about the reason of why deploying the selection method and the math formulas for the channel selection. Arounf 1/3-1/2 page.
Channel selection module identifies and retains a fixed subset of the top-$K$ most discriminative channels based on the statistical divergence of their spectral entropy between seizure and non-seizure states. For each channel $c$ in the training set, we first compute the Power Spectral Density (PSD), donated as $P_c(f)$, using the Fast Fourier Transform over a 0.5-70 Hz frequency band. The PSD is normalized to form a probability distribution $\hat{P}_c(f)$. The spectral entropy $H_c$ is then donated as:
\begin{equation}
    H_c = - \sum_{f} \hat{P}_c(f) \log_2 \hat{P}_c(f)
    \label{eq:spc_entr}
\end{equation}
To determine which channels are the most informative, we compute a discriminative score $S_c$ for each channel. This score quantifies the separation power of the channel by comparing the distribution of its entropy values across the seizure and non-seizure labels. Specifically, we calculate the means ($\mu_{c, \text{seizure}}, \mu_{c, \text{normal}}$) and variance ($\sigma^2_{c, \text{seizure}}, \sigma^2_{c, \text{normal}}$) of the spectral entropy for each channel across the training set. 
The discriminative score $S_c$ is then calculated by a variance-pooled effect size metric: 
\begin{equation}
    S_c = \frac{\left| \mu_{c, \text{seizure}} - \mu_{c, \text{normal}} \right|}{\sqrt{\frac{1}{2}\left(\sigma^2_{c, \text{seizure}} + \sigma^2_{c, \text{normal}}\right)}},
    \label{eq:ranking_score}
\end{equation}
Channels are ranked in order of $S_c$ and We effectively apply a global mask to retain only the top-$K$ channels with the highest scores. This channel selection ensures minimizing inference latency by reducing the visual token budget and directs the encoding focus exclusively toward the most informative channels, thereby facilitating robust detection across heterogeneous EEG inputs.

\subsection{Canvas of Neuron Signal}\label{cns_module}

Direct tokenization of multi-channel signal like EEG ($X \in \mathbb{R}^{C \times T}$) represents a scaling challenge. Using numeric embedding layers for LLMs often result in a sequence length that is very large \cite{liu2025picture}. Using waveform plots as an input works in some cases \cite{liu2025picture} but the input image is inherently sparse with large amounts of white space.
Furthermore, as number of channels increases, the overlapping amplitudes make it difficult to distinguish distinct channel.
As an alternative to sparse time-domain visualization, frequency domain via Mel Spectrogram construction was proposed for robust audio encoding \cite{radford2023robust}.
While this approach aims to feed a more rich time-frequency representation to the model, it still has some constraints.
Firstly, the approach introduces computational overheads of converting time-series data to time-frequency ($\mathcal{O}(T \log T)$). Secondly, it still does not solve the issue of feeding multiple channels of data to a LLM or VLLM in an efficient manner. In this method we will need $C$ images for $C$ channels.

Thus, our approach of using \texttt{CNS} attempts to address this issue. The intensity maps will operate in the time domain and will stack multiple channels together, ensuring compact signal representations. Each pixel value will thus directly represent the normalized signal activity.

\textbf{Normalization and clipping.}
Clipping of the EEG signals will allow us to supress outliers while normalizing it will stabilize the amplitude distribution. 
Given a EEG clip $X$, we apply amplitude clipping and normalization with a fixed bound:
\begin{equation}
\tilde{X}_{c,t}=\mathrm{clip}(X_{c,t},-A,A),\qquad \\
Z_{c,t}=\frac{\tilde{X}_{c,t}+A}{2A}\in[0,1]
\label{eq:clip_norm}
\end{equation}
Here, $Z\in[0,1]^{C\times T}$ is a bounded representation whose values can be mapped to pixel intensities.

\textbf{Intensity-map construction.}
We convert $Z$ into an image where the vertical axis corresponds to channels and the horizontal axis corresponds to time. We first get the grayscale encoding using:
\begin{equation}
    P_{c,t}=\left\lfloor 255\cdot Z_{c,t}\right\rceil \in \{0,\dots,255\}
    \label{eq:gray_enc}
\end{equation}
To exploit the “third-dimension” capacity of images, we then map each scalar intensity to an RGB triplet via a colormap $g:[0,1]\rightarrow[0,1]^3$:
\begin{equation}
    I_{c,t,:}=g(Z_{c,t})\in[0,1]^3,
    \label{color_pi}
\end{equation}

yielding a compact image $I\in\mathbb{R}^{C\times T\times 3}$. 
% Unlike waveform plots where color is typically used only to distinguish lines (and thus carries limited quantitative meaning), this representation uses pixel intensity (and optionally color) to encode signal magnitude, increasing information density within a compact image \cite{}.
To represent the compactness of the information, we define the Information Density ($\rho$) as the ratio of non-zero feature pixels to the total pixel area $H \times W$. In waveform plots, $\rho \ll 0.1$ due to the inherent sparsity of line drawings. In contrast, our Intensity Map achieves $\rho \approx 1.0$. Thus, allowing for a $C$-fold increase in channel capacity without expanding the model's input dimensionality.

% @Yan Chen
% @Jie do we need empirical calculation for Information density

\subsection{Network Architecture.} The intensity map from our novel \texttt{CNS} module is encoded by the vision encoder which is then trained using instruction tuning according to the system prompt in Appendix \ref{prmt}. The frozen LLM decoder then predicts the seizure status.

% \subsection{Multimodality Foundation Vision Model}
% \subsection{Model Structure}
% This part mainly discuss about the mechanism of VLLM. Should include the core info that why VLLM can correctly capture the feature in our intensity map. Not very sure how exactly this part will be written. Jie may give us some guidance. Around half page

% move this to the vegining of the section

% \begin{figure}[htpt]
% \centering
% \centerline{\includegraphics[width=\columnwidth]{fig_model_structure.pdf}}
% \caption{Proposed model for seizure detection that uses intensity representation}
% \label{fig:Network}
% \end{figure}

\section{Experiments}
\subsection{Experiment Setup}
% This part discuss about the the info of dataset and metrics we are using. We need to emphysis that we are using binary f1 and using this is reasonable for 

We evaluated NeuroCanvas on Widely used scalp EEG-seizure dataset: TUSZ \cite{shah2018temple} and CHB-MIT \cite{guttag2010chb}. The details of preprocessing and the datasets is in Appendix \ref{data_pre}.

\textbf{Metrics.} To show model's performance under extreme class imbalance scenario, we report accuracy, precision, recall and binary F1-score \cite{tholke2023class}. Notably, we prioritize the binary F1-score over the weighted F1-score often cited in prior literature \cite{afzal2024rest,tang2021self,zhang2022epileptic}. Since the weighted F1-score is heavily dominated by the majority class, and this can mask poor performance on the minority seizure class.

% @Yan Chen
% Put Preprocessing and Dataset split: in appendix

\textbf{Baselines.} We compared NeuroCanvas with SOTA seizure detection models, including REST \cite{afzal2024rest} and DCRNN \cite{tang2021self}. Except these latest models, we also included more traditional used deep learning models which are also widely used for seizure detection: ResNet-LSTM \cite{lee2022real}, CNN-LSTM \cite{ahmedt2020neural}, Transformer \cite{vaswani2017attention}, and GRU \cite{cho2014learning}. Additionally, we included a text-augmented time series tasks finetuned LLM: Time-LLM \cite{jin2023time} to compare our model with the numeric input LLM. To ensure the rigor and validity of our comparative analysis, we restricted our baselines to methods which are replicable.

\textbf{Model training.} We implemented NeuroCanvas using the Qwen2.5-VL-7B-Instruct architecture as the backbone foundation model \cite{bai2025qwen2}. To adapt the visual-linguistic capabilities of the pre-trained model to the domain of EEG signal analysis, we adopted a fine-tuning strategy: the weights of the LLM decoder were frozen, while the vision encoder and the cross-modal projector (merger) were kept trainable. Optimization was performed using AdamW with a cosine decay learning rate scheduler. The training process utilized a per-device batch size of 2 with 8 gradient accumulation steps and was executed with BF16 precision on 2 NVIDIA H100 GPUs. More details in Appendix.

% \[
% \mathrm{F1}=\frac{2\cdot \mathrm{Precision}\cdot \mathrm{Recall}}{\mathrm{Precision}+\mathrm{Recall}}
% =\frac{2\mathrm{TP}}{2\mathrm{TP}+\mathrm{FP}+\mathrm{FN}}
% \]

% This part we will describe the EEG preprocessing method. Including the filtering, clip and labeling method. Around 7 rows.

% \subsection{Implementation details}

% \subsection{Result of channel selection}

% \subsection{Baselines}
% We compared NeuroCanvas with cutting-edge seizure detection models, including REST \cite{afzal2024rest} and DCRNN \cite{tang2021self}. Except these latest models, we also included more traditional used deep learning models which are also widely used for seizure detection: ResNet-LSTM \cite{lee2022real}, CNN-LSTM \cite{ahmedt2020neural}, Transformer \cite{vaswani2017attention}, and GRU \cite{cho2014learning}. Additionally, we included a text-augmented time series tasks finetuned LLM: Time-LLM to compare our model with the numeric input LLM.

% Find out their F1 calculation methods, if same, use their result and mark in the table. Use a barplot to compare ourmodel with Evobrain

% Also put parameters in the table
\begin{table}[ht!]
  \centering
  \caption{Seizure detection performance and efficiency on TUSZ dataset.}
  \label{tab:tusz_result}
  % \small
  \begin{small}
  \resizebox{0.9\linewidth}{!}{
    \begin{tabular}{l|cccc|cc}
      \toprule
      %\multirow{2}{*}{\textbf{Model}} 
      & \multicolumn{4}{c|}{\textbf{Model Performance}} &
      \multicolumn{2}{c}{\textbf{Model Efficiency}} \\
      \cline{2-7}
      \textbf{Model}
      & Accuracy & Precision & Recall & Binary F1 &
      Inference time (ms) & Parameter Count\\
      \midrule

      GRU & $0.6434 \pm 0.0521$ & $0.1793 \pm 0.0211$ & $0.6969 \pm 0.0596$ & $0.2845 \pm 0.0273$ & $3.9701 \pm 0.0406$ & $763K$ \\
      CNN-LSTM & $0.6528 \pm 0.4327$ & $0.1207 \pm 0.0194$ & {\boldmath $0.9297 \pm 0.0504$} & $0.2128 \pm 0.0295$ & $4.0706 \pm 0.0344$ & $349K$ \\
      Transformer & $0.7966 \pm 0.0128$ & $0.2702 \pm 0.0138$ & $0.5896 \pm 0.0272$ & $0.3702 \pm 0.0141$ & $3.918 \pm 0.0014$ & $490K$\\
      ResNet-LSTM & $0.7349 \pm 0.0448$ & $0.2564 \pm 0.0327$ & $0.8326 \pm 0.0198$ & $0.3910 \pm 0.0359$ & $2.9263 \pm 0.0211$ & $3M$ \\
      \midrule
      DCRNN & $0.7905 \pm 0.0588$ & $0.2643 \pm 0.0463$ & $0.8376 \pm 0.0431$ & $0.4074 \pm 0.0546$ & $7.8728 \pm 0.0086$ & $150K$ \\
      REST & $0.7457 \pm 0.0358$ & $0.2418 \pm 0.0166$ & $0.7687 \pm 0.0355$ & $0.3643 \pm 0.0150$ & $1.3480 \pm 0.0004$ & $10K$ \\
      \midrule
      Time-LLM & $0.4944 \pm 0.1608$ & $0.1595 \pm 0.0358$ & $0.8805 \pm 0.0217$ & $0.2686 \pm 0.0517$ & $92.031 \pm 10.440$ & $1.1B$\\
      \midrule
      Our Model (19Ch DV) & $0.7961 \pm 0.0358$ & $0.3068 \pm 0.0124$ & $0.8080 \pm 0.0225$ & $0.4438 \pm 0.0137$ & $1054.3 \pm 23.71$ & $7B$ \\
      Our Model (19Ch \texttt{CNS}) & {\boldmath $0.8306 \pm 0.0212$} & {\boldmath $0.3580 \pm 0.0061$} & $0.8225 \pm 0.0324$ & {\boldmath $0.5022 \pm 0.0119$} & $126.75 \pm 1.251$ & $7B$ \\
      Our Model (8Ch \texttt{CNS}) & $0.7984 \pm 0.0183$ & $0.3182 \pm 0.0083$ & $0.7544 \pm 0.0416$ & $0.4505 \pm 0.0142$ & $50.112 \pm 0.8413$ & $7B$ \\
      Our model (4Ch \texttt{CNS}) &$0.7542 \pm 0.0141$ & $0.2734 \pm 0.0092$ & $0.7973 \pm 0.0391$ & $0.4059 \pm 0.0125$ & $32.171 \pm 0.6931$ & $7B$ \\
      Our model (2Ch \texttt{CNS}) & $0.7139 \pm 0.0098$ & $0.2414 \pm 0.0107$ & $0.7880 \pm 0.0231$ & $0.3686 \pm 0.0104$ & $19.324 \pm 0.6577$ & $7B$ \\      
      \bottomrule
    \end{tabular}
    }
    \end{small}
  % \vspace{-2mmA}
\end{table}

\begin{table}[t]
  \centering
  \caption{Seizure detection performance and efficiency on CHB-MIT dataset.}
  \label{tab:chbmit_result}
  % \small
  % \setlength{\tabcolsep}{3pt}
  %\renewcommand{\arraystretch}{1.15}

  \resizebox{0.9\linewidth}{!}{%
    \begin{tabular}{l|cccc|cc}
      \toprule
      %\multirow{\textbf{Model}} 
      & \multicolumn{4}{c|}{\textbf{Model Performance}} &
      \multicolumn{2}{c}{\textbf{Model Efficiency}} \\
      \cline{2-7}
      \textbf{Model}
      & Accuracy & Precision & Recall & Binary F1 &
      Inference time (ms) & Parameter Count\\
      \midrule      

      GRU & $0.6818 \pm 0.1057$ & $0.1830 \pm 0.0401$ & $0.7967 \pm 0.0586$ & $0.2951 \pm 0.0501$ & $3.8462 \pm 0.0151$ & $763K$ \\
      CNN-LSTM & $0.7505 \pm 0.0834$ & $0.2410 \pm 0.0328$ & {\boldmath $0.8817 \pm 0.0174$} & $0.3742 \pm 0.0450$ & $4.4932 \pm 0.0891$ & $349K$ \\
      Transformer & $0.7606 \pm 0.0195$ & $0.2173 \pm 0.0113$ & $0.6450 \pm 0.0569$ & $0.3318 \pm 0.0151$ & $4.2726 \pm 0.0262$ & $490K$\\
      ResNet-LSTM & $0.7583 \pm 0.0134$ & $0.2145 \pm 0.0046$ & $0.7561 \pm 0.0047$ & $0.3342 \pm 0.0051$ & $2.2490 \pm 0.0753$ & $3M$ \\
      \midrule
      DCRNN & $0.8574 \pm 0.005$ & $0.3288 \pm 0.008$ & $0.7462 \pm 0.002$ & $0.4564 \pm 0.008$ & $0.9004 \pm 0.1812$ & $150K$ \\
      REST & $0.6623 \pm 0.1312$ & $0.1658 \pm 0.0492$ & $0.7028 \pm 0.1522$ & $0.2608 \pm 0.0471$ & $19.736 \pm 6.0947$ & $10K$ \\
      \midrule
      Time-LLM & $0.6838 \pm 0.0392$ & $0.0993 \pm 0.0056$ & $0.3622 \pm 0.0385$ & $0.1555 \pm 0.0040$ & $34.496 \pm 1.2309$ & $1.1B$\\
      \midrule
      Our Model (16Ch DV) & $0.8710 \pm 0.0084$ & $0.3636 \pm0.0039$ & $0.6612 \pm 0.0091$ & $0.4692 \pm 0.0059$ & $752.44 \pm 4.012$ & $7B$ \\
      Our Model (16Ch \texttt{CNS}) & $0.8885 \pm 0.0072$ & $0.3954 \pm0.0056$ & $0.7372 \pm 0.0103$ & $0.5146 \pm 0.0059$ & $107.75 \pm 1.114$ & $7B$ \\
      Our Model (8Ch \texttt{CNS}) & {\boldmath $0.9337 \pm 0.0130$} & {\boldmath $0.6270 \pm0.0067$} & $0.4282 \pm 0.0142$ & $0.5089 \pm 0.092$ & $57.91 \pm 0.9934$ & $7B$\\
      Our model (4Ch \texttt{CNS}) & $0.8965 \pm 0.0207$ & $0.4183 \pm0.0042$ & $0.7425 \pm 0.0183$ & {\boldmath $0.5351 \pm 0.092$} & $31.65 \pm 0.7826$ & $7B$ \\
      Our model (2Ch \texttt{CNS}) & $0.8487 \pm 0.0286$ & $0.2884 \pm0.0075$ & $0.6043 \pm 0.0204$  & $0.3905 \pm 0.123$ & $19.41 \pm 0.6137$ & $7B$ \\      
      \bottomrule
    \end{tabular}%
  }

  \vspace{-2mm}
\end{table}

\subsection{Superior Performance}
% We evaluate NeuroCanvas on the TUSZ and CHB-MIT datasets and compared it against baseline models in Tables \ref{tab:tusz_result} \& \ref{tab:chbmit_result}.
\textit{TUSZ (Table \ref{tab:tusz_result}):} We observe improvements of \{0.4944 - 0.8306, 0.1207 - 0.3580, 0.2128 - 0.5022\} in accuracy, precision, and binary F1 scores respectively. These improvements underscore the increased seizure detection accuracy of our model. The slightly lower recall score compared to baseline is mitigated by the fact that models with higher recall (CNN-LSTM) have very poor precision. Thus, our model has better precision-recall balance. \textit{CHB-MIT(Table \ref{tab:chbmit_result}):} The improvements of the model are also showcased in the CHB-MIT dataset with improvements of \{0.6623 - 0.9337, 0.0993 - 0.6270, 0.3622 - 0.7425, 0.1555 - 0.5351\} in accuracy, precision, and binary F1 scores respectively. We can see the same pattern for this dataset as well: baseline models with higher recall have very poor precision while our model maintains a better precision-recall balance.

% whereas the best baseline (DCRNN) reaches only 0.4074 F1. This shows \textit{more than 20\%} relative improvement in F1, pointing to a significant boost in detection accuracy. 
% NeuroCanvas also outperforms other models like ResNet-LSTM (F1 0.3910) and REST (F1 0.3643) by a large margin, setting a new standard on TUSZ. Importantly, the F1 gain comes from a much better precision-recall balance. Our model achieves 35.8\% precision at 82.3\% recall, which greatly reduces false alarms compared to previous methods, such as DCRNN's 26.4\% precision at 83.8\% recall. This improved balance means NeuroCanvas detects seizures as reliably as the top RNN-based models while producing far fewer false positives, which is essential for practical use.

\subsection{Indepth Model Analysis}

\textbf{Importance of \texttt{CNS}.}
We encode the raw waveforms into tokens in the Time-LLM framework\cite{jin2023time}. 
Using \texttt{CNS} image representations, instead of encoded raw waveforms \cite{jin2023time} lead to an improvement of binary F1 score of \textbf{~87\%} and \textbf{~231\%} in TUSZ and CHB-MIT dataset respectively.
This shows that the \texttt{CNS} encoding offers a rich, informative visual input that pretrained vision models can easily understand, unlike token sequences.
Furthermore, the proposed \texttt{CNS} image representation produces much better results than a direct time-series image representation of the EEG. 
Compared to using time-series image approach  \cite{liu2025picture}, our model achieved a relative improvement of \textbf{~13\%} and \textbf{~10\%} in TUSZ and CHB-MIT datasets respectively. This highlights the effectiveness of using \texttt{CNS} instead of a lineplot as an input to VLLM.

\begin{table}[ht!]
  \caption{Ablation experiment results on the TUSZ dataset.}
  \label{tab:ab_ch_select}
  \centering
  \small
  \resizebox{0.78\linewidth}{!}{%
  \begin{tabular}{l|cccc}
    \toprule
    \textbf{Model} & \textbf{Acc.} & \textbf{Prec.} & \textbf{Rec.} & \textbf{F1} \\
    \midrule
    NeuroCanvas (8 highest ch, HT) &$0.7984 \pm 0.0183$ & $0.3182 \pm 0.0083$ & $0.7544 \pm 0.0416$ & $0.4505 \pm 0.0142$ \\
    NeuroCanvas (8 lowest ch, HT) &$0.7705 \pm 0.0129$ & $0.2990 \pm 0.0048$ & $0.8319 \pm 0.0732$ & $0.4353 \pm 0.0126$ \\
    \midrule
    NeuroCanvas (19Ch HT, qwen2.5-7B) &$0.8306 \pm 0.0212$ & $0.3580 \pm 0.0061$ & $0.8225 \pm 0.0324$ & $0.5022 \pm 0.0119$ \\
    NeuroCanvas (19Ch HT, qwen2.5-3B) &$0.8287 \pm 0.0153$ & $0.3473 \pm 0.0054$ & $0.7909 \pm 0.0218$ & $0.4827 \pm 0.0096$ \\
    \bottomrule
  \end{tabular}}
\end{table}

% \begin{wrapfigure}[28]{r}{0.51\textwidth}
%   \centering
%   \includegraphics[width=\linewidth]{fig1_overview.pdf}
%   \caption{Overview of the NeuroCanvas framework.}
%   \label{fig:fr_overview}
% \end{wrapfigure}
% Attention analysis around here

\begin{wrapfigure}[18]{r}{0.56\columnwidth}
\centering
\includegraphics[width=\linewidth]{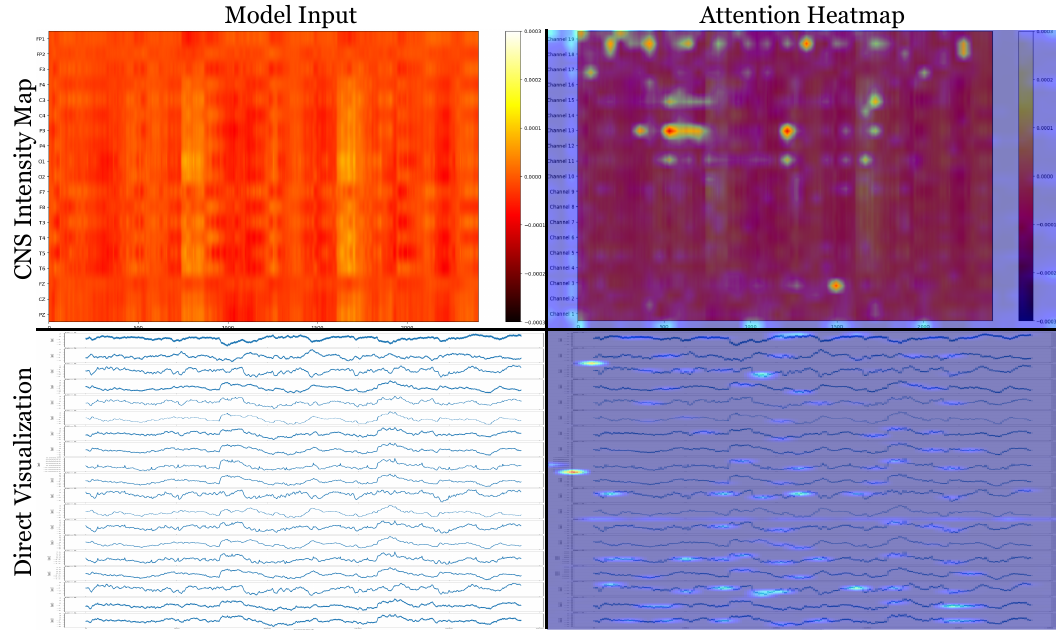}
\caption{Attention heatmap on \texttt{CNS} intensity map and direct time series visualization figure.}
\label{fig:attn_map}
\end{wrapfigure}

\begin{figure}[t]
    \centering
    \includegraphics[width=0.6\linewidth,height=0.6\textheight,keepaspectratio]{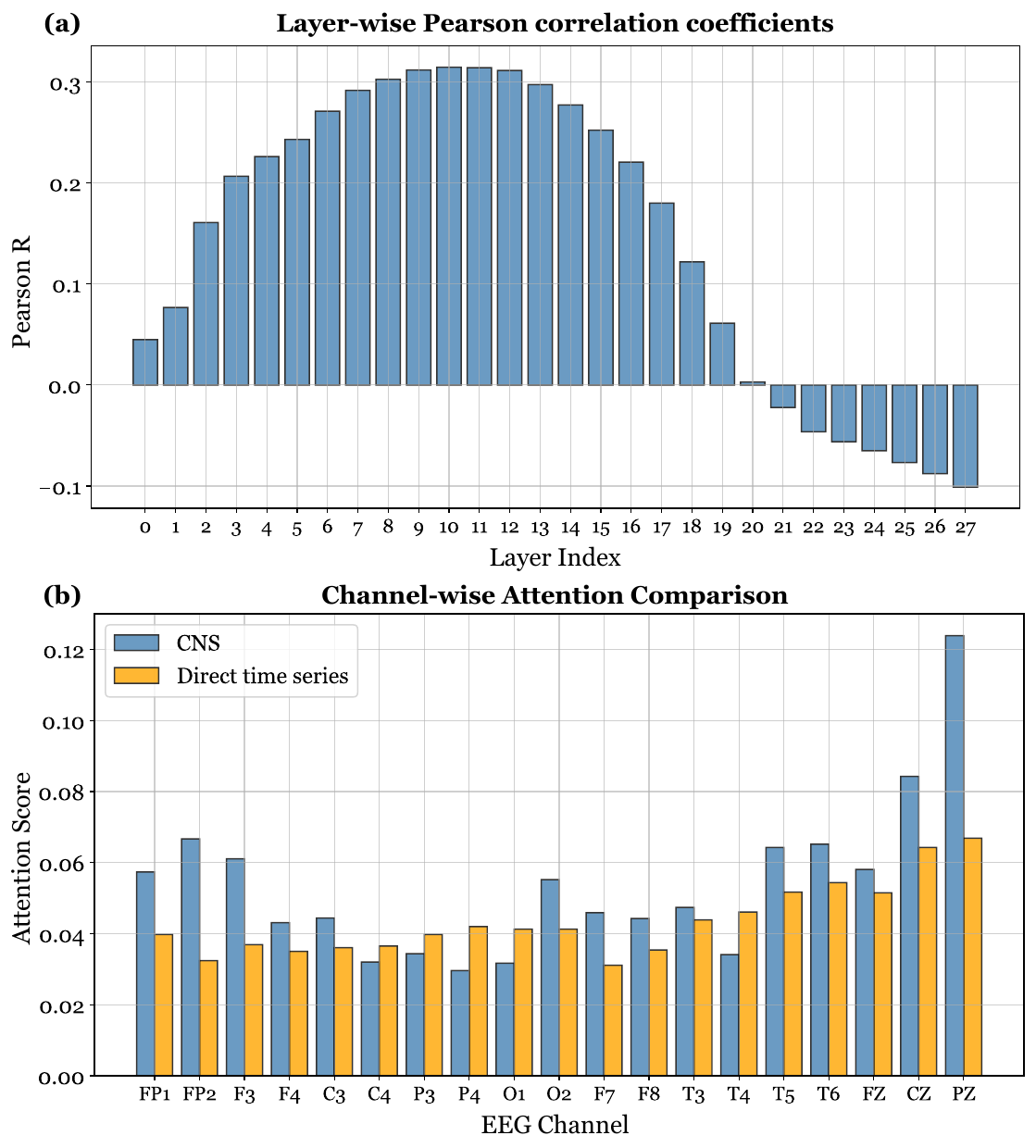}
    \caption{\textbf{(a)} Layer-wise Pearson correlation coefficients between "\texttt{CNS}" and Direct Time-series. \textbf{(b)} Comparison of average attention scores across 19 EEG channels.}
    \label{fig:pearsonr_attn_score}
\end{figure}
\FloatBarrier

\textbf{Effect of \texttt{ECS}.}
In realistic scenarios, not all EEG channels may be available or helpful. Thus, we tested our model with fewer channels, using the \texttt{ECS} to choose the most informative electrodes. With only the top 8 channels, our model achieved binary F1 score higher than feeding lineplot to the VLLM. 
Even with 2 channels, our method achieved \textbf{~73\%} of the peak binary F1 score. Thus showcasing that carefully selected channels can be used to effectively detect seizure.
On top of that, using just 4 channels even gave the best result in CHB-MIT. This outcome may be attributed to the bipolar EEG montage employed in the dataset, which emphasizes localized differential activity and reduces common-mode noise, thereby allowing a smaller set of channels to remain highly discriminative. 

\textbf{Model Efficiency.} 
Even with over six times more parameters, our model achieved better inference time than Time-LLM at 8ch (TUSZ) and 4ch (CHB-MIT). Despite Time-LLM using patch tokenizer to encode the input signals, \texttt{CNS} and \texttt{ECS} allowed our model to outperform it.

\textbf{Attention heatmap visualization.} To validate that our our \texttt{CNS} enhances the model's ability to more efficiently capture useful information in the EEG representation, we visualize the attention score heatmap maps of the VLLM's vision encoder (Figure \ref{fig:attn_map}). In the direct visualization (right), the visual attention weights are sparse and frequently scattered across non-informative background regions, as circled in the red blanket (Bad attention). We attribute this failure to the inherent visual sparsity of waveform line plots, normal Vision Transformers processes images via patch embeddings, which struggles to extract robust semantic features from patches dominated by whitespace with only minor line signals \cite{dosovitskiy2020image}. The lack of textural coherency prevents the model from anchoring its attention on the physiological signal, leading to stochastic focus. In contrast, the \texttt{CNS} representation (left) effectively reformulates the seizure detection task into a visual object recognition problem. By encoding signal amplitude into chromatic intensity, \texttt{CNS} transforms transient seizure events into salient spatiotemporal representations. As observed in the heatmap, the VLLM's visual attention is accurately directed toward signal regions strongly correlated with seizure detection.

\begin{wrapfigure}[18]{l}{0.46\columnwidth}
\centering
\captionsetup{skip=3pt}
\includegraphics[width=\linewidth,height=0.28\textheight,keepaspectratio]{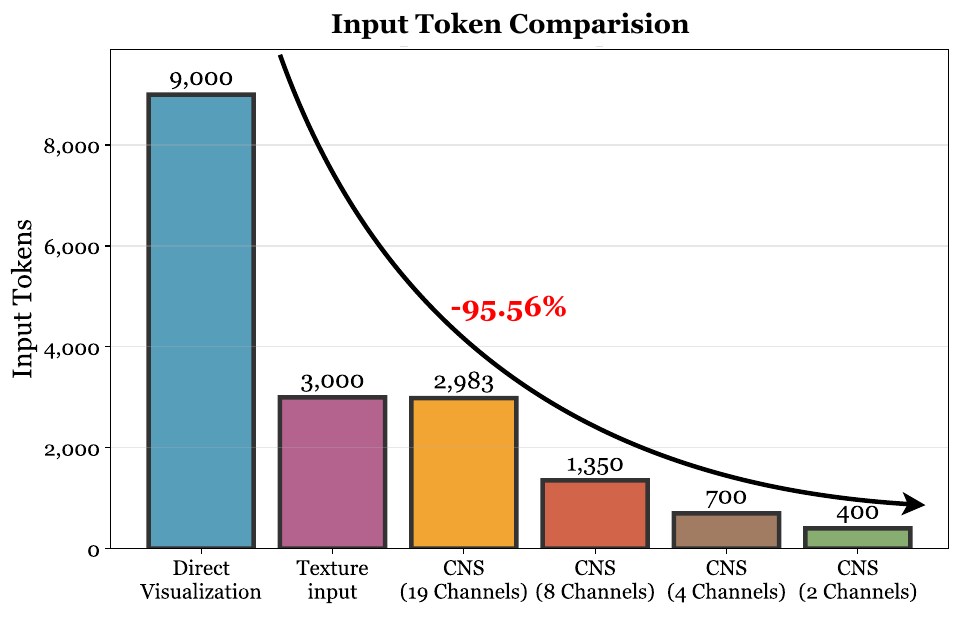}
\caption{Number of input tokens across EEG representation pipelines. Compared with numeric and direct visualization inputs, \texttt{CNS} achieves up to a 95.56\% reduction in input tokens.}
\label{fig:token_compr}
\end{wrapfigure}
% \vspace{-6mm}

\textbf{Layer-wise Pearson correlation analysis.} To find out where along the vision encoder the two EEG representation methods begin to induce qualitatively different channel evidence, we compute a layer-wise Pearson $R$ bar plot between \texttt{CNS} and direct visualization (Figure \ref{fig:pearsonr_attn_score}a). The resulting trend provides evidence that \texttt{CNS} changes not only what regions are attended, but also when representation-dependent channel prioritization emerges in the encoder hierarchy. Specifically, the positive correlations in earlier layers suggest that both renderings share a common, representation agnostic stage where the model recovers coarse channel localization and allocates attention in broadly consistent ways. However, later layers show near-zero or negative correlations, indicating a re-ranking of channel importance where CNS and visualization inputs emphasize different channels, consistent with Figure \ref{fig:attn_map}.
% However, the subsequent collapse toward near-zero and negative correlations in later layers indicates a systematic re-ranking of channel importance: at decision-forming depths, \texttt{CNS} and direct visualization inputs drive the model to rely on different channel subsets, which aligning with the different attention weights shown in Figure \ref{fig:attn_map}. 
This layer-dependent divergence is consistent with the general principle that earlier layers tend to encode more generic, transferable features, whereas deeper layers become increasingly specialized to input statistics and task-relevant abstractions \cite{li2024surprising,raghu2021vision}. Importantly, the negative tail bars argues against interpreting \texttt{CNS} as a cosmetic reparameterization. Instead, it functions as an inductive bias that reorganizes how channel evidence is composed downstream. While attention-based explanations must be interpreted cautiously, attention flow is explicitly designed to trace how saliency propagates through transformer depth \cite{abnar2020quantifying}, making the observed late-layer inversion a meaningful signal of representational bifurcation rather than a visualization artifact.

\textbf{Attention score analysis.} We further quantify the attention distribution on a channel-level. Figure \ref{fig:pearsonr_attn_score}(b) shows that \texttt{CNS} produces a consistently higher and more discriminative attention allocation in most of the 19 channels. This indicates that \texttt{CNS} helps the model spend its limited visual capacity on channel-discriminative evidence rather than distributing attention uniformly over visually redundant regions. Additionally, visual attention of \texttt{CNS} concentrates on a smaller subset of channels (e.g. FP1, FP2), and this concentration indicates improved channel selectivity, suggesting that the model is less likely to diffuse its capacity across weakly informative channels and more likely to leverage discriminative channels. From a neuro signal perspective, this behavior is also consistent with the fact that seizure events can behave in spatially non-uniform patterns across channels, where only part of the image provides strong discriminative evidence at a certain time \cite{fisher2017operational}.

\textbf{Token compression efficiency.} Beyond improving attention allocation, \texttt{CNS} substantially reduces the effective input length presented to the VLLM. As shown in Figure \ref{fig:token_compr} direct visualization consumes $9{,}000$ input visual tokens due to  over large, mostly redundant canvases. In contrast, \texttt{CNS} compresses the input by converting the multi-channel window into a compact, information-dense image, reducing token usage to $\sim3{,}000$ for 19 channels and further to 1,350/700/400 tokens when retaining 8/4/2 channels, respectively up to a 95.56\% input token reduction.

% \textbf{The attention distribution analysis for \texttt{CNS}.}
% To further investigate how \texttt{CNS} contribute the prediction, we analysis how the attention weight on each channel for the\texttt{CNS} module that convert the EEG to intensity map and the Direct Visualization method that convert the EEG to individual sub-figures.
% In Figure~\ref{fig:compr_tok} shows, ``TP v.s. TP'' shows the \texttt{CNS} consistly have higher attention score than the Direct Visualization method, inditate it improves the signal-to-background ratio, stabilizing the attention allocation without shifting the underlying channel preference.
% For the ``TP v.s. FN'', \texttt{CNS} allocate lower attention weight for some channels to achive the right prediction, indicating its ability in avoiding failure modes when waveform plots indice attention toward unhelpful regions.
% For the ``FN v.s. TP'', the \texttt{CNS} fails when it fail in capture important channel like $FP1$, indicating for sample depends on waveform morphology, the critical information is compressed by the intensity map, which indicating the further invesigation for how to highlight important channels in the intensity map.
% For the ``FN v.s. FN'', the attention weight pattern is similar to ``TP v.s. TP'', indicating the reason may comes from the data distribution or label noise.
\begin{figure}[H]
\centering
\includegraphics[width=\linewidth,height=0.75\textheight,keepaspectratio]{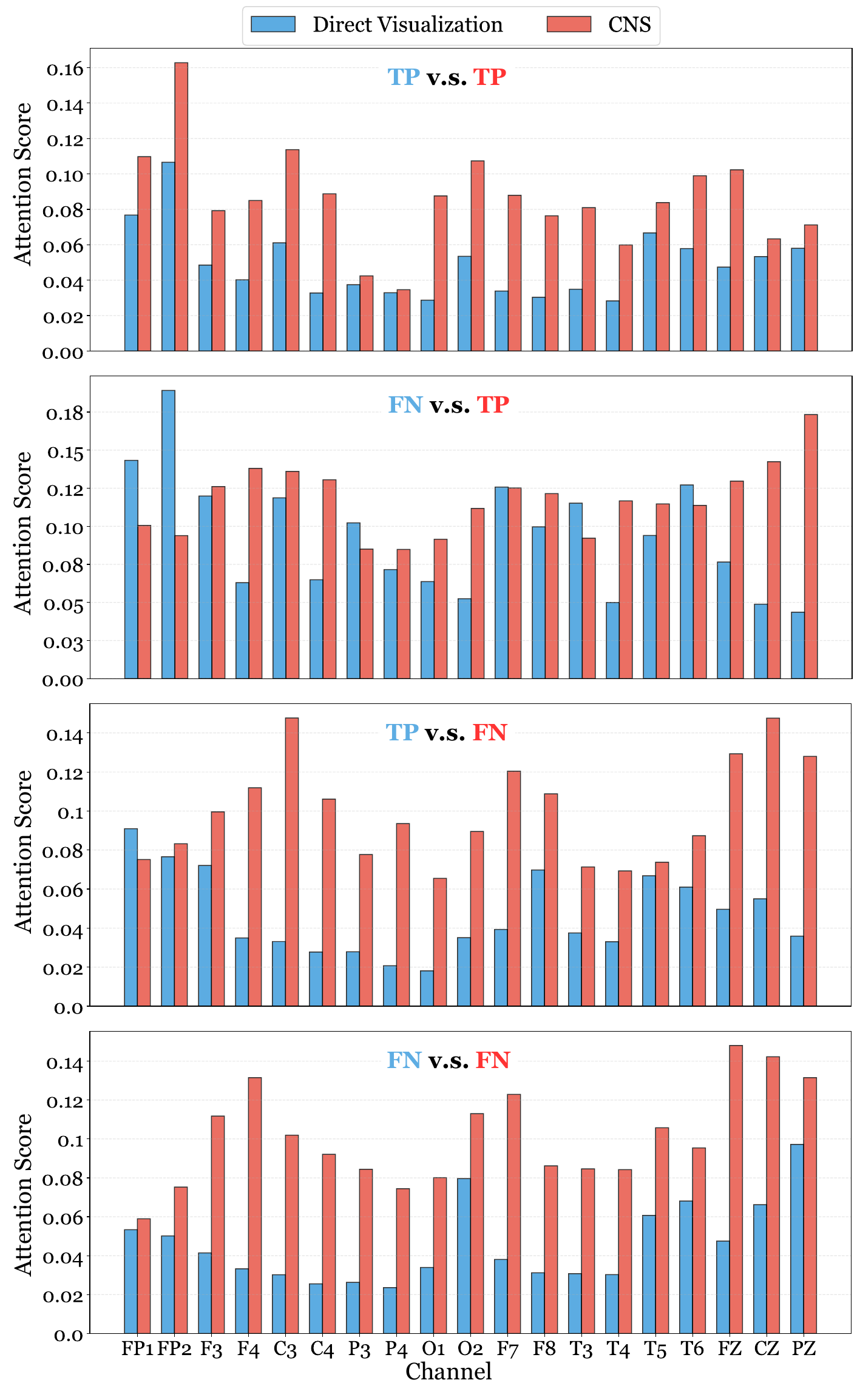}
\caption{Why \texttt{CNS} helps or fails, the attention distribution analysis. The ``TP'' denotes the true positive, and the ``FN'' means the false positive. The \textcolor{blue}{blue} represents the Direct Visualization, and the \textcolor{red}{red} denotes \texttt{CNS}.}
\label{fig:case_result}
\end{figure}
\textbf{Attention distribution analysis for \texttt{CNS}.}

To better understand how the \texttt{CNS} module contributes to predictions, we analyze the attention weights assigned to each EEG channel by \texttt{CNS}, which converts EEG signals into intensity maps, and compare them to the Direct Visualization method, which converts EEG signals into individual sub-figures. As shown in Figure~\ref{fig:case_result}, the analysis is conducted across four scenarios: ``TP v.s. TP'', ``TP v.s. FN'', ``FN v.s. TP'', and ``FN v.s. FN''.
\ding{182} \textbf{TP v.s. TP}: \texttt{CNS} consistently achieves higher attention scores than the Direct Visualization method. This indicates that \texttt{CNS} improves the signal-to-background ratio, stabilizing attention allocation while maintaining the underlying channel preference.
\ding{183} \textbf{TP v.s. FN}: In cases where \texttt{CNS} correctly predicts while the Direct Visualization method fails, \texttt{CNS} allocates lower attention weights to specific channels, enabling it to avoid failure modes caused by waveform plots inducing attention toward irrelevant regions.
\ding{184} \textbf{FN v.s. TP}: \texttt{CNS} struggles to capture critical channels, such as $FP1$, in waveform morphology-dependent samples. This suggests that, for samples relying heavily on waveform shape, the intensity map compresses essential information. Further work is needed to enhance the representation of such channels within the intensity map.
\ding{185} \textbf{FN v.s. FN}: The attention weight distribution for this scenario resembles the pattern observed in "TP v.s. TP." This behavior may be attributed to issues in data distribution or label noise, which warrant additional investigation.
Overall, this analysis highlights the strengths of \texttt{CNS} in improving attention allocation and avoiding failure modes, while identifying limitations in handling waveform-dependent samples. Future research should explore strategies to better preserve critical channel information in intensity maps.

\subsection{Ablation Experiment}

% \begin{table}[ht!]
%   \caption{Ablation experiment for the base model.}
%   \label{tab:ab_base_model}
%   \centering
%   \small
%   \resizebox{\linewidth}{!}{%
%   \begin{tabular}{l|cccc}
%     \toprule
%     \textbf{Model} & \textbf{Acc.} & \textbf{Prec.} & \textbf{Rec.} & \textbf{F1} \\
%     \midrule
%     NeuroCanvas (19Ch HT, qwen2.5-7B) &$0.8306 \pm 0.0212$ & $0.3580 \pm 0.0061$ & $0.8225 \pm 0.0324$ & $0.5022 \pm 0.0119$ \\
%     NeuroCanvas (19Ch HT, qwen2.5-3B) &$0.8287 \pm 0.0153$ & $0.3473 \pm 0.0054$ & $0.7909 \pm 0.0218$ & $0.4827 \pm 0.0096$ \\
%     \bottomrule
%   \end{tabular}}
% \end{table}

\textbf{Effectiveness of Entropy-Guided Channel Selection.} To verify the effectiveness of \texttt{ECS}, we compared our proposed strategy (selecting the Top-$8$ channels with the \textit{highest} discriminative spectral entropy scores) against a counter-factual baseline that selects the $8$ channels with the \textit{lowest} scores.  As shown in Table~\ref{tab:ab_ch_select}, the model employing the Top-$8$ channels with the highest spectral entropy secures superior performance, achieving relative improvements of \textbf{3.62\%} in accuracy and \textbf{3.49\%} in binary F1-score. This result indicates that spectral entropy acts as an effective metric for channel discriminability. Prioritizing high-entropy channels preserves more informative spatiotemporal dynamics for \texttt{CNS}, which translates into better overall detection quality. Notably, using low spectral entropy score channels attains higher recall but but substantially lower precision, suggesting that selecting low-entropy channels biases the model toward a more permissive decision boundary and increases false alarms.

\textbf{Impact of Base Model.} We further investigate the impact of the base model by deploying NeuroCanvas with Qwen2.5-VL-7B and Qwen2.5-VL-3B under the same 19-channel \texttt{CNS} setting (Table \ref{tab:ab_ch_select}). Both settings achieve comparable accuracy, indicating that our \texttt{CNS} representation are not strongly reliant on model scale. Nevertheless, the 7B model consistently provides stronger precision  and a higher binary F1 score, reflecting improved calibration and fewer false positives at similar recall. These findings suggest that while a larger base model relatively elevates the performance, the performance improvements from NeuroCanvas are largely driven by the proposed \texttt{CNS} and will remain robust even when the base model is downsized.

% \subsection{EEG encoding Representation Analysis}
% Put some other input representaions to show that ours are the most efficient

% \subsubsection{Hyperparameter Studies}

% \subsubsection{VLLM Variants Analysis}

% \section{Case study: When CNS Helps and When it Fails}

% Case Study to be included

\section{Conclusion and Future Work}
In this work, we introduced NeuroCanvas, a novel framework detecting seizure by transforming multi-channel EEG signals into information-dense intensity map visual representations tailored for VLLMs. Our approach includes two key components: \texttt{CNS} and \texttt{ECS} to address the critical challenges of multi-channel heterogeneity and computational inefficiency in seizure detection. Extensive experiments on the TUSZ and CHB-MIT datasets demonstrate the superiority of NeuroCanvas. Our model achieves a binary F1-score of 0.5022 on the challenging TUSZ benchmark, outperforming SOTA baselines by over 20\%. It also reduces  88\% of inference latency compare to traditional EEG visual representations. Future work will focus on optimizing NeuroCanvas for deployment on resource-constrained medical devices. Additionally, we plan to leverage reinforcement learning to further enhance the model's reasoning capabilities and diagnostic precision.

% Authors are kindly asked to make their submissions as accessible as possible
% for everyone including people with disabilities and sensory or neurological
% differences. Tips of how to achieve this and what to pay attention to will be
% provided on the conference website \url{http://icml.cc/}.

% \section*{Software and Data}

% If a paper is accepted, we strongly encourage the publication of software and
% data with the camera-ready version of the paper whenever appropriate. This can
% be done by including a URL in the camera-ready copy. However, \textbf{do not}
% include URLs that reveal your institution or identity in your submission for
% review. Instead, provide an anonymous URL or upload the material as
% ``Supplementary Material'' into the OpenReview reviewing system. Note that
% reviewers are not required to look at this material when writing their review.

% \section*{Acknowledgements}

% \textbf{Do not} include acknowledgements in the initial version of the paper

% If a paper is accepted, the final camera-ready version can (and usually should)
% include acknowledgements.  Such acknowledgements should be placed at the end of
% the section, in an unnumbered section that does not count towards the paper
% page limit. Typically, this will include thanks to reviewers who gave useful
% comments, to colleagues who contributed to the ideas, and to funding agencies
% and corporate sponsors that provided financial support.

\FloatBarrier
\clearpage
\section*{Impact Statement}

The datasets utilized in this research, the Temple University Hospital Seizure Corpus (TUSZ) and the CHB-MIT Scalp EEG Database, are anonymized and publicly accessible resources that adhere to ethical standards for patient privacy \cite{shah2018temple,guttag2010chb}. The authors declare no conflicts of interest, and the methodology presented does not generate harmful insights. NeuroCanvas demonstrates a significant reduction in inference latency and improved performance in the seizure detection domain. This research highlights the potential of foundation models to democratize access to high-quality neurological monitoring across diverse clinical settings.

\section*{Acknowledgement}
This research was partially funded by the National Institutes of Health (NIH) under award 1R01EB03710101. The views and conclusions contained in this document are those of the authors and should not be interpreted as representing the official policies, either expressed or implied, of the NIH.

% Authors are \textbf{required} to include a statement of the potential broader
% impact of their work, including its ethical aspects and future societal
% consequences. This statement should be in an unnumbered section at the end of
% the paper (co-located with Acknowledgements -- the two may appear in either
% order, but both must be before References), and does not count toward the paper
% page limit. In many cases, where the ethical impacts and expected societal
% implications are those that are well established when advancing the field of
% Machine Learning, substantial discussion is not required, and a simple
% statement such as the following will suffice:

% ``This paper presents work whose goal is to advance the field of Machine
% Learning. There are many potential societal consequences of our work, none
% which we feel must be specifically highlighted here.''

% The above statement can be used verbatim in such cases, but we encourage
% authors to think about whether there is content which does warrant further
% discussion, as this statement will be apparent if the paper is later flagged
% for ethics review.

\nocite{langley00}

\FloatBarrier
\bibliography{example_paper,ref}
\bibliographystyle{icml2026}

%%%%%%%%%%%%%%%%%%%%%%%%%%%%%%%%%%%%%%%%%%%%%%%%%%%%%%%%%%%%%%%%%%%%%%%%%%%%%%%
%%%%%%%%%%%%%%%%%%%%%%%%%%%%%%%%%%%%%%%%%%%%%%%%%%%%%%%%%%%%%%%%%%%%%%%%%%%%%%%
% APPENDIX
%%%%%%%%%%%%%%%%%%%%%%%%%%%%%%%%%%%%%%%%%%%%%%%%%%%%%%%%%%%%%%%%%%%%%%%%%%%%%%%
%%%%%%%%%%%%%%%%%%%%%%%%%%%%%%%%%%%%%%%%%%%%%%%%%%%%%%%%%%%%%%%%%%%%%%%%%%%%%%%
\clearpage
\appendix
\section{Entropy scores of channels}

\begin{figure}[ht!]
\centering
\centerline{\includegraphics{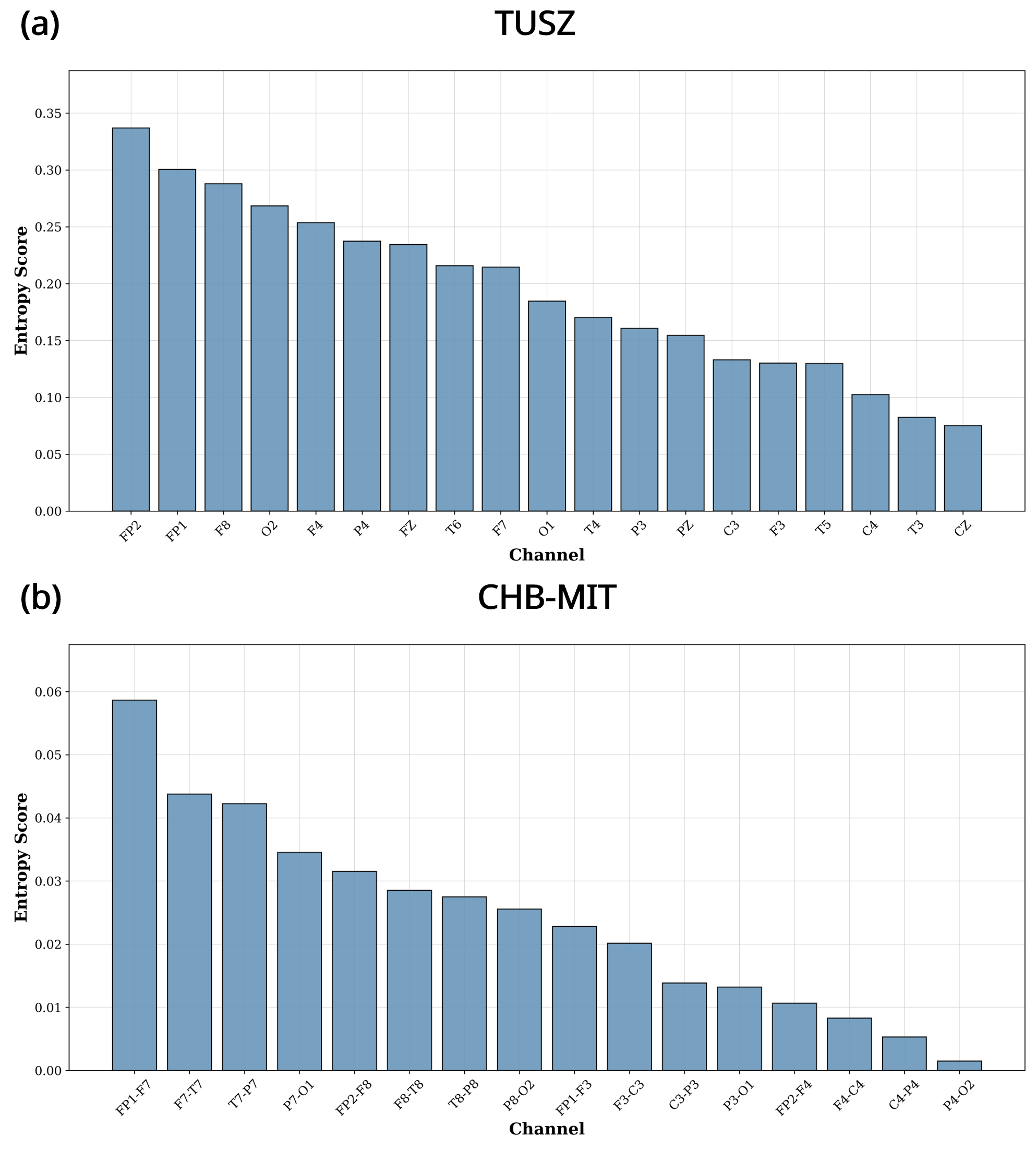}}
\caption{Entropy scores calculted for (a) TUSZ (b) CHB-MIT dataset}
\label{fig:ent_scores}
\end{figure}

Figure \ref{fig:ent_scores} shows the entropy scores calculated for TUSZ and CHB-MIT dataset. Interestingly, the frontopolar channels (FP1, FP2) are quite discriminative for TUSZ. On the other hand, the bipolar recordings of CHB-MIT show that "FP1-F7" and "F7-T7" are the most discriminative.

% \section{Implementation of Time-LLM}

% \section{CNS Maps}

\section{Model Architecture} \label{moedl_arc}
For different visual representations, they have different resolutions. In direct visualization, multichannel EEG is rendered as waveform plot. To preserve readability, the input image is scaled with the number of channels and the desired amplitude. In \texttt{CNS} canvas, since the amplitude of the signal has been encoded as color, the resolution of the \texttt{CNS} canvas is only related to the number of input channels. Detailed resolutions of each types of visual representation are in Appendix (Table \ref{tab:reso}). 

\begin{table}[h]
    \centering
    \caption{Resolution comparison between Direct Visualization and \texttt{CNS} across different channel settings.}
    \label{tab:reso}
    \begin{tabular}{lc}
        \toprule
        Visual Representation & Resolution (Width $\times$ Height) \\
        \midrule
        \multicolumn{2}{c}{\textbf{TUSZ}} \\
        \midrule
        Direct Visualize 19 Channels & $1538 \times 4650$ \\
        \texttt{CNS} 19 Channels              & $1464 \times 860$  \\
        \texttt{CNS} 8 Channels               & $1464 \times 379$  \\
        \texttt{CNS} 4 Channels               & $1464 \times 204$  \\
        \texttt{CNS} 2 Channels               & $1464 \times 111$  \\
        \midrule
        \multicolumn{2}{c}{\textbf{CHB-MIT}} \\
        \midrule
        Direct Visualize 16 Channels & $1538 \times 3914$ \\
        \texttt{CNS} 16 Channels              & $1464 \times 729$  \\
        \texttt{CNS} 8 Channels               & $1464 \times 379$  \\
        \texttt{CNS} 4 Channels               & $1464 \times 204$  \\
        \texttt{CNS} 2 Channels               & $1464 \times 111$  \\
        \bottomrule
    \end{tabular}
\end{table}

NeuroCanvas is deployed on a pretrained base VLLM with three standard blocks: ($i$) a \emph{vision encoder} $E_v$ that converts an input image into a sequence of visual tokens, ($ii$) a \emph{cross-modal projector} $P$ that maps visual features into the LLM hidden space, and ($iii$) a \emph{LLM decoder} $\mathcal{M}$ that performs conditional generation given visual and text tokens. Given a canvas image $I_c$, the vision encoder extracts spatial embeddings
$Z_v = E_v(I_v)$, which are then projected into the language space $\tilde{Z}_v = P(Z_v)$. In parallel, a task prompt $p$ (Details in Appendix \ref{prmt}) is tokenized and embedded as $Z_t = \mathrm{Emb}(p)$. The decoder then consumes the concatenation of cross-modal tokens and predicts the output.

\section{Prompt for training} \label{prmt}
We cast seizure detection as prompt-guided binary decision. Concretely, the model is prompted to generate a short answer token from a constrained label set $\{\texttt{Seizure},\texttt{Non-seizure}\}$).

Example Prompt: "Question: Input image is a 19-Channel EEG signal intensity map, the color refers to the amplitude of the EEG, is the image represents Seizure or Non-seizure? \textbackslash n Options: \textbackslash n 1. Seizure \textbackslash n 2. Non-seizure".

\section{Details of Preprocessing and Dataset Splitting} \label{data_pre}

\textbf{TUSZ.} In v2.0.3 release, TUSZ contains 315 subjects and 1643 recording sessions. All recordings utilize 19 channels in standard 10-20 system. Importantly, TUSZ reflects real clinical imbalance: the corpus includes over 504 hours of annotated EEG, with seizure activity comprising about 36 hours (around 7\%)
\textbf{CHB-MIT.} This dataset contains recordings from 22 subjects, each having between 9 and 42 sessions. Due to inconsistencies in the available channels for some subjects, we utilized the 16 channels in standard 10-20 system that are present in all subjects. 

\textbf{Preprocessing.} To stay consistent with previous studies \cite{tang2021self,afzal2024rest,zhang2022epileptic}, we resampled the EEG signals into 200Hz for TUSZ dataset and 256Hz for CHB-MIT dataset. The continuous EEG recordings are then segmented into non-overlapping clips with a fixed duration of $T=12$ seconds. We selected this 12-second window size based on prior empirical benchmarks, which demonstrate that it provides the optimal trade-off between capturing sufficient temporal context for seizure detection and latency for the model inference \cite{tang2021self,afzal2024rest}.

\textbf{Dataset splitting.} For the TUSZ dataset, we followed the officially defined data partitions. The original provided training set was randomly split into training and validation subsets with a 90/10 ratio, while the official evaluation set served as the standardized test set. For the CHB-MIT dataset, since no predefined splits are provided, we randomly divided the data into approximately 80\% for training, 10\% for validation, and 10\% for testing. To ensure robust evaluation, this splitting was performed on a subject level, preventing the model from being tested on patients included in the training set. Detailed statistics regarding the data distribution across splits are provided in the Appendix. For training set of both TUSZ and CHB-MIT, we employed random under-sampling on the non-seizure class so that the number of non-seizure clips matched the number of seizure clips. The test sets remained imbalanced to reflect real imbalanced scenarios.

\section{Computing Layer-wise Pearson Correlations }
For each layer $\ell$, we derive an attention-induced saliency over visual tokens using attention rollout up to $\ell$, then project token scores back to a two-dimensional map using the model’s image placeholder mask and the image grid metadata. We subsequently aggregate the saliency map into an attention vector by averaging within each channel’s spatial band. Finally, we compute Pearson’s $r$ between the two attention vectors (\texttt{CNS} vs. direct visualization) per layer and report the mean across paired samples. 

\section{Further Analysis of Case Study}

\begin{figure}[!htbp]
\centering
\centerline{\includegraphics[width=\linewidth,height=0.9\textheight,keepaspectratio]{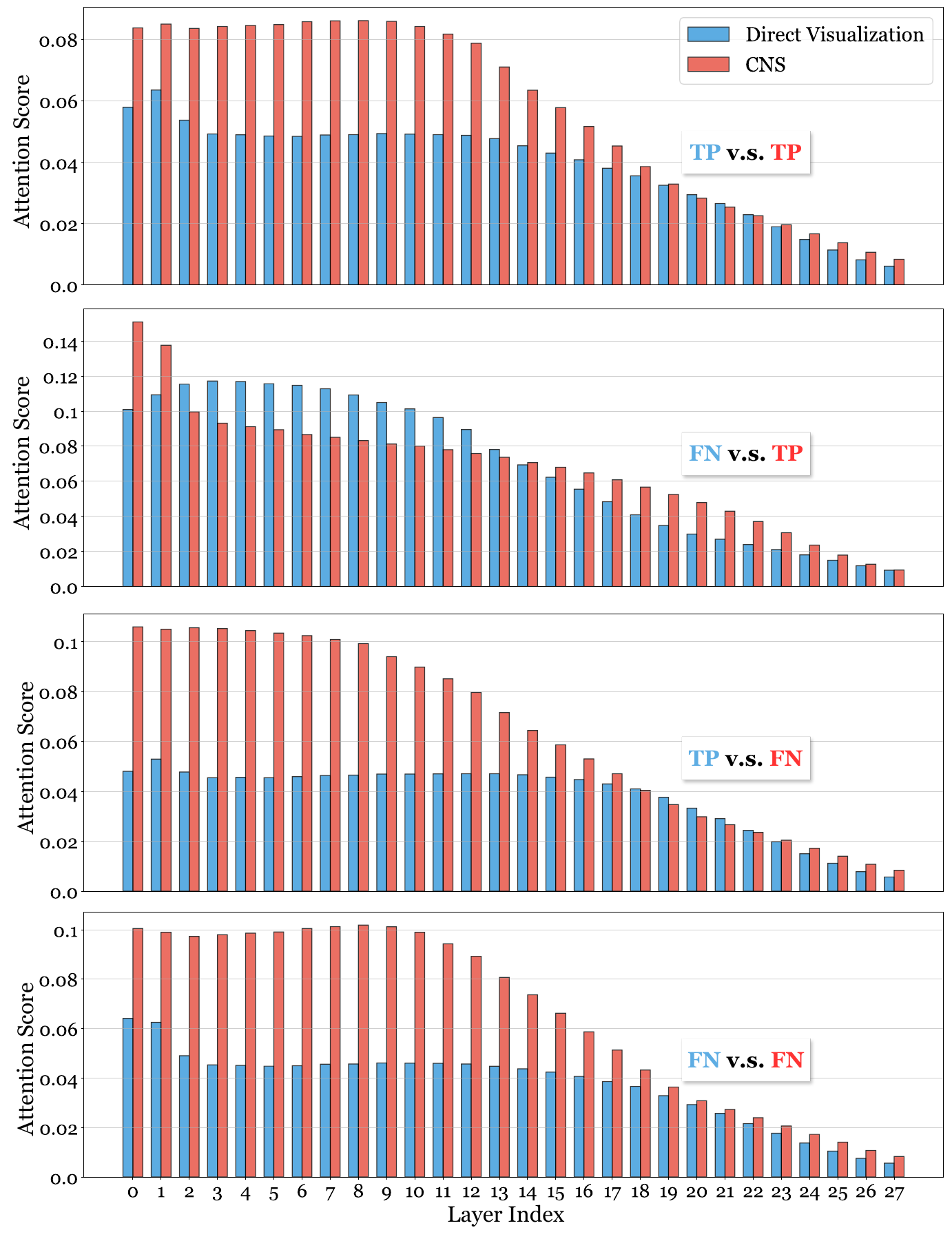}}
\caption{Why \texttt{CNS} helps or fails, the attention distribution analysis in Layer-wise. The ``TP'' denotes the true positive, and the ``FN'' means the false positive. The \textcolor{blue}{blue} represents the Direct Visualization, and the \textcolor{red}{red} denotes \texttt{CNS}.}
\label{fig:layerwise_attn}
\end{figure}

To better understand how \texttt{CNS} changes model behavior beyond aggregate metrics, we conduct a case study over four outcome conditions defined by the detection correctness under \texttt{CNS} versus direct visualization: (1) TP/TP, (2) TP/FN, (3) FN/FN, and (4) FN/TP. To ensure statistical representativeness, we randomly sampled 100 instances for each outcome condition and conducted the analysis based on their averaged attention profiles. For each condition, we inspected the mean attention score per layer (averaged across channels), and the attention distribution on each channel. We use these analyses as a diagnostic lens to characterize the routing of representation-dependent evidence through the vision encoder. 

\textbf{Condition 1: TP/TP.} In condition where both models correctly detect the seizure, the channel-wise attention distribution is strongly aligned (Pearson's $r= 0.732$, $p < 0.001$). This correlation indicates that both representations prioritize a similar subset of channels; however, \texttt{CNS} consistently assigns higher attention magnitudes across most channels. Layer-wise analysis corroborates this, as \texttt{CNS} maintains a higher mean attention score, whereas the direct pipeline remains lower throughout. The combination of high cross-representation alignment and stronger attention suggests that in this condition, \texttt{CNS} acts as an amplifier rather than altering the fundamental evidence path. It does not change what evidence is used when the waveform already exposes discriminative cues, but rather increases the saliency of that shared evidence, making it accessible earlier in the hierarchy. Practically, this implies that \texttt{CNS} improves the signal-to-background ratio, stabilizing attention allocation without shifting the underlying channel preference that drives correct decisions.

\textbf{Condition 2: TP/FN.} In condition where \texttt{CNS} succeeds but the direct visualization fails is characterized by a systematic disagreement in channel prioritization (Pearson's $r= -0.405$, $p = 0.085$). The two representations induce partially inverted channel-wise attention distribution, indicating \texttt{CNS} effectively performs evidence re-ranking. Notably, the direct visualization allocates substantial attention mass to a broad set of channels, which often emphasizes posterior or midline regions. However, it fails to detect the seizure. In contrast, \texttt{CNS} corrects this representation-induced bias by concentrating attention on a distinct subset of channels. Layer-wise analysis show that direct visualization maintains relatively high mean attention across many mid-layers. This supports that direct visualization method faces misallocated search without correctly deetcting the seizure. On the other hand, \texttt{CNS} exhibits very strong early-layer attention followed by a rapid decline, consistent with the earlier localization of discriminative evidence. This indicates that \texttt{CNS} can avoid failure modes where waveform plots induce attention toward visually salient but diagnostically unhelpful structures.

\textbf{Condition 3: FN/TP.} In condition where \texttt{CNS} fails while the direct visualization succeeds, the channel profiles are nearly uncorrelated (Pearson's $r= -0.064$, $p = 0.796$), indicating their reliance on fundamentally different channel preferences. Although \texttt{CNS} again shows substantially higher mean attention across early layers, this increased focus does not translate into a correct detection, demonstrating that greater attention magnitude is not equivalent to better evidence quality. According to the distribution of attention  on channels, the \texttt{CNS} strongly allocates attention to central channels, while the direct visualization places relatively more attention on front channels. This divergence suggests that the discriminative evidence for these samples depends on waveform morphology (e.g. transient sharpness or brief rhythmic evolution), that remains visible in line plots but becomes distorted during amplitude-to-color encoding. 

\textbf{Condition 4:FN/FN.} In condition where both \texttt{CNS} and direct visualization fail to detect the seizure, channel-wise attention shows relatively strong agreement between \texttt{CNS} and direct visualization (Pearson's $r= 0.418$, $p = 0.075$). While both representations shows weak concentration on certain channels, \texttt{CNS} spreads high attention across a broad range of electrodes. Layer-wise, \texttt{CNS} maintains consistent high attention, whereas direct visualization remains lower and flatter.
This pattern reflects a condition that neither representation provides sufficiently separable evidence within the signal clip. This might be caused by signal-noise ratio, weak seizure morphology, or strong non-seizure confounds. In this context, attention fails to concentrate on a discriminative channel subset. The broader and higher attention score observed in \texttt{CNS} is consistent with the model trying hard but failing to resolve a clear decision boundary, highlighting that representation improvements alone cannot compensate for segments with intrinsically low evidence or label noise.

%%%%%%%%%%%%%%%%%%%%%%%%%%%%%%%%%%%%%%%%%%%%%%%%%%%%%%%%%%%%%%%%%%%%%%%%%%%%%%%
%%%%%%%%%%%%%%%%%%%%%%%%%%%%%%%%%%%%%%%%%%%%%%%%%%%%%%%%%%%%%%%%%%%%%%%%%%%%%%%

\end{document}